\documentclass[letterpaper,10pt,conference]{ieeeconf}
\IEEEoverridecommandlockouts
\usepackage{cite}
\usepackage{amsmath,amssymb,amsfonts}
\usepackage{graphicx}
\usepackage{textcomp}
\usepackage{xcolor}
\usepackage{booktabs}
\usepackage{multirow}
\usepackage[hidelinks]{hyperref}

\renewcommand{\rq}[1]{{\color{black}#1}}
\providecommand{\rv}[1]{#1}
\renewcommand{\rv}[1]{{\color{black}#1}}

\newcommand{\NAME}{WEVO}

\makeatletter
\AtBeginDocument{%
  \setlength{\floatsep}{0.55\baselineskip plus 0.1\baselineskip minus 0.15\baselineskip}%
  \setlength{\textfloatsep}{0.9\baselineskip plus 0.1\baselineskip minus 0.2\baselineskip}%
  \setlength{\dblfloatsep}{0.55\baselineskip plus 0.1\baselineskip minus 0.15\baselineskip}%
  \setlength{\dbltextfloatsep}{0.9\baselineskip plus 0.1\baselineskip minus 0.2\baselineskip}%
  \setlength{\intextsep}{0.55\baselineskip plus 0.1\baselineskip minus 0.15\baselineskip}%
  \setlength{\abovecaptionskip}{0.25\baselineskip}%
  \setlength{\abovedisplayskip}{0.45\baselineskip plus 1pt minus 2pt}%
  \setlength{\belowdisplayskip}{0.45\baselineskip plus 1pt minus 2pt}%
  \setlength{\abovedisplayshortskip}{0.2\baselineskip plus 1pt minus 1pt}%
  \setlength{\belowdisplayshortskip}{0.3\baselineskip plus 1pt minus 1pt}%
  \setlength{\jot}{2pt}%
}
\def\section{\@startsection{section}{1}{\z@}{0.9ex plus 0.8ex minus 0.3ex}%
  {0.4ex plus 0.4ex minus 0ex}{\normalfont\normalsize\centering\scshape}}%
\def\subsection{\@startsection{subsection}{2}{\z@}{0.9ex plus 0.8ex minus 0.3ex}%
  {0.4ex plus 0.3ex minus 0ex}{\normalfont\normalsize\itshape}}%
\makeatother
\def\BibTeX{{\rm B\kern-.05em{\sc i\kern-.025em b}\kern-.08em
    T\kern-.1667em\lower.7ex\hbox{E}\kern-.125emX}}

\begin{document}


\title{\LARGE \bf WEVO: \underline{W}orld Model-\underline{E}nabled \underline{V}LA \underline{O}ptimization for Whole-Body Loco-Manipulation}


\author{%
Qianzhong Chen$^{1,2*}$\quad
Muyuan Lin$^{1}$\quad
Yuyin Sun$^{1}$\\
Mac Schwager$^{2}$\quad
Cheng-Hao Kuo$^{1}$\quad
Roberto Mart\'{\i}n-Mart\'{\i}n$^{1,3}$\\[0.6em]
{\normalsize $^{1}$Amazon\qquad
$^{2}$Stanford University\qquad
$^{3}$The University of Texas at Austin}%
\thanks{$^{*}$Work done during an internship at Amazon.}%
}


\maketitle
\thispagestyle{empty}
\pagestyle{empty}

\begin{abstract}
Reinforcement learning (RL) can improve vision-language-action (VLA) policies beyond imitation, but humanoid exploration is unsafe and simulator RL can exploit residual sim-to-real gaps. We present \NAME{}, a teleoperation-free framework for post-training a whole-body humanoid VLA in an action-conditioned video world model (WM). Simulated manipulation and locomotion play data improve contact-dynamics and viewpoint-change prediction. A VLA trained on synthetic demonstrations collects $20$ autonomous real robot rollouts per task to align the WM with deployment observations. The VLA, whole-body controller (WBC), and video WM then form a closed-loop RL environment with dense progress rewards. Across three real robot tasks, play data improves every evaluated WM metric and downstream policy success; video WM RL raises mean success from $23.3\%$ to $75.0\%$, compared with $48.3\%$ for real robot RL. \rq{We therefore leverage the complementary strengths of both environments: simulation provides policy initialization and broad coverage, while the video WM offers a deployment-aligned environment for policy improvement.} 
Project website: \texttt{https://qianzhong-chen.github.io/wevo.github.io/}.
\end{abstract}

\section{Introduction}
Vision-language-action (VLA) models~\cite{rt2,openvla,pi0,pi05,gr00t} couple pretrained vision-language backbones with action heads learned from demonstrations, whose scale and quality strongly determine performance. Collecting such demonstrations through teleoperation is particularly difficult for high-degree-of-freedom (DoF) humanoids. Whole-body controllers (WBCs)~\cite{sonic,twist2} coordinate balance, locomotion,  and bimanual control but do not remove this data bottleneck. Real-to-sim-to-real (R2S2R) systems based on 3D Gaussian Splatting (3DGS)~\cite{legs,vlk,3dgs} instead reconstruct deployment scenes and generate whole-body trajectories without teleoperation. Policies trained on these synthetic demonstrations can transfer to hardware, but residual appearance and dynamics gaps still require learning from deployment experience.

\begin{figure}[t!]
  \centering
  \includegraphics[width=\columnwidth]{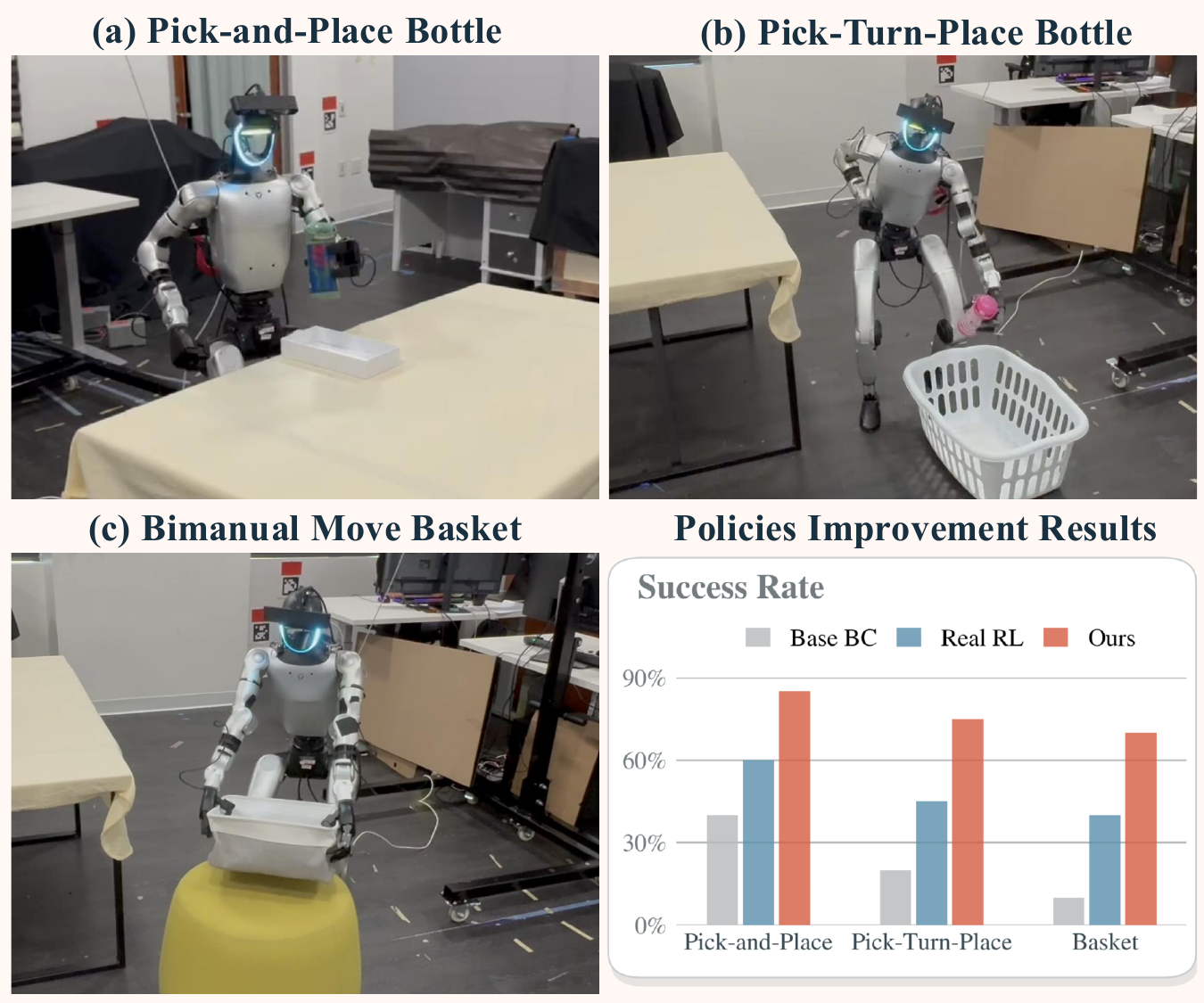}
  \caption{\textbf{Real robot tasks and policy improvement.} We evaluate three tasks, Pick-and-Place Bottle, Pick-Turn-Place Bottle, and Bimanual Move Basket. The bottom right panel compares the base BC policy, real robot RL, and our video WM RL, which performs best on all three tasks.}
  \label{fig:pull}
  \vspace{-0.55\baselineskip}
\end{figure}

Reinforcement learning can improve a behavior-cloned (BC) policy under compounding errors and residual sim-to-real mismatch, but VLA post-training requires long-horizon rewards and optimization through diffusion or flow action heads. Offline-to-online methods often update simple MLP policies~\cite{hu2023imitation,yang2024robot,huang2024mentor}, limiting expressiveness through their tractable action densities. Advantage-weighted methods support richer policies~\cite{wu2025robocopilot,intelligence2025pi,peng2019advantage, chen2025conrft}, but often require human DAgger interventions. RL-100~\cite{lei2025rl} differentiates the PPO objective~\cite{schulman2017proximal} through the full denoising process at high computational cost. Steering~\cite{wagenmaker2025steering}, residual methods~\cite{xiao2025self,ankile2025residual}, and DICE-RL~\cite{sun2026prior} adapt pretrained policy outputs more directly. Nevertheless, these approaches rely on continued real robot interaction or fixed offline data, neither of which supports exploratory whole-body humanoid learning.

Learned \emph{world models} (WMs) shift policy optimization away from hardware by predicting future observations under candidate actions. Existing systems use either compact dynamics representations or generative video models. Latent dynamics models and explicit 3D digital twins~\cite{dreamer,wu2022daydreamer,drema,diwa} enable sample-efficient learning, but are usually scene-specific and do not provide the open-world visual inputs required by VLAs. Generative video WMs instead adapt large pretrained models to robot actions. VLAW~\cite{vlaw} treats successful WM rollouts as behavior-cloning targets, while WMPO~\cite{wmpo} and related methods optimize policies on generated trajectories~\cite{world4rl,rise,worldgymnast,worldvlaloop}. WoVR~\cite{wovr}, WorldSample~\cite{worldsample}, and PlayWorld~\cite{playworld} improve rollout initialization, transition selection, and autonomous data collection, respectively. However, these methods remain limited to fixed-base tabletop manipulation and use human teleoperation data during training.

\begin{figure*}[h!]
  \centering
  \includegraphics[width=\textwidth]{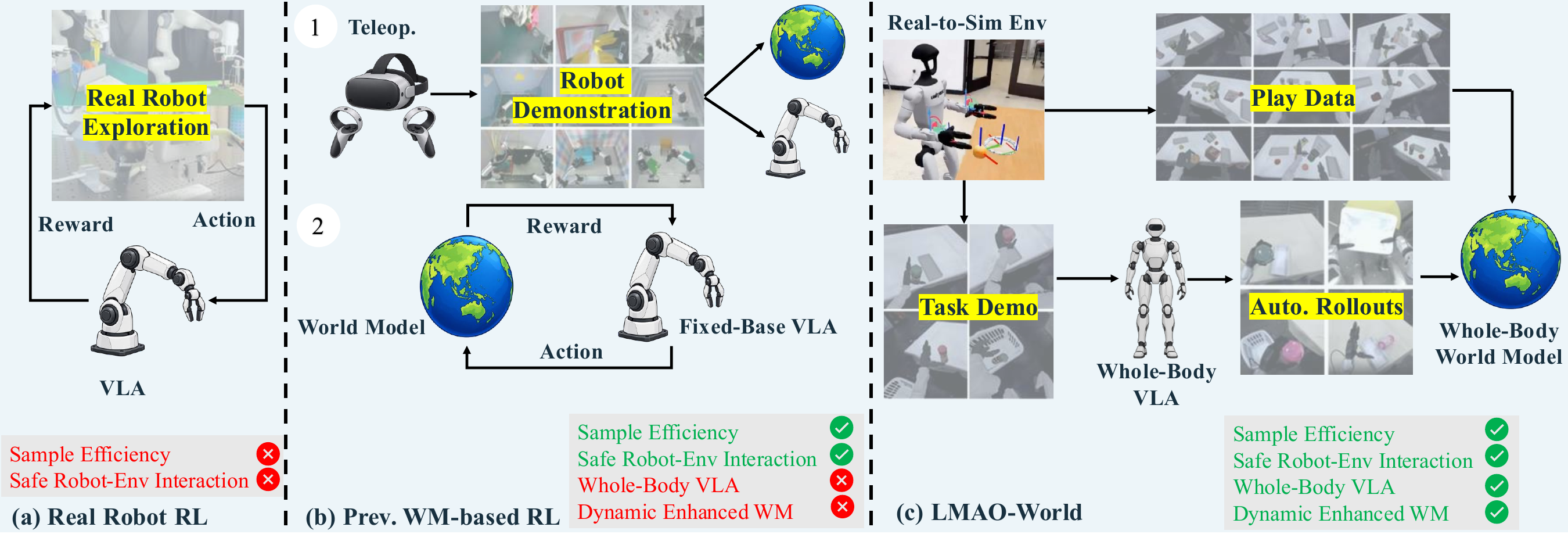}
  \caption{\textbf{\NAME{}: teleoperation-free world model RL for humanoid loco-manipulation.} (a) Hardware RL is costly and risky. (b) Prior WM-RL focuses on fixed-base tasks and uses teleoperation data. (c) \NAME{} combines synthetic demonstrations, simulated play data, and autonomous real robot rollouts for whole-body policy improvement.}
  \label{fig:teaser}
  \vspace{-0.35\baselineskip}
\end{figure*}

\rq{These limitations raise a central question: how should scalable synthetic data and limited autonomous policy rollouts be combined for whole-body humanoid VLA post-training without teleoperation? \NAME{} uses simulation to train the initial VLA and broaden the WM's dynamics coverage. Autonomous real robot rollouts then close the remaining visual sim-to-real gap in its predictions, after which the video WM serves as the policy-improvement environment.} This forms the flywheel shown in Fig.~\ref{fig:teaser}. Our contributions are:
\vspace{-0.2\baselineskip}
\begin{enumerate}
\setlength{\topsep}{0.2\baselineskip}
\setlength{\itemsep}{0pt}
\setlength{\parsep}{0pt}
\rq{
\item We build \NAME{}, to our knowledge the first WM-RL system for whole-body humanoid VLAs beyond fixed-base tabletop manipulation, coupling egocentric video generation with whole-body dynamics and dense progress rewards.
\item We develop autonomous whole-body play data for WM mid-training: manipulation play improves contact-dynamics prediction, while locomotion play improves consistency under large egocentric viewpoint changes.
\item We study how photorealistic simulation and a video WM can form a teleoperation-free policy-improvement flywheel. Experiments on three loco-manipulation tasks show that simulation provides policy initialization and broad coverage, while the video WM provides a more useful environment for policy improvement.
}
\end{enumerate}
\vspace{-0.35\baselineskip}
\rq{We design the system around two practical constraints: unconstrained autonomous play is unsafe on hardware, while RL can exploit residual visual sim-to-real gap in even a photorealistic 3DGS-based simulator. We therefore assign complementary roles to the two environments. The simulator supplies policy initialization and physically grounded play data; autonomous real robot rollouts align the video WM with deployment observations, making it a more faithful environment for policy improvement.}

\section{Background}
\subsection{The LEGS Simulator}
\label{sec:legs}
The LEGS simulator~\cite{legs} is an R2S2R data engine for the Unitree G1. It combines a 3DGS reconstruction of the static scene with interactable object meshes recovered using SAM 3D~\cite{sam3d}. MuJoCo~\cite{mujoco} models robot--object contacts, while the 3DGS scene provides background rendering. SONIC~\cite{sonic} tracks scripted locomotion and manipulation primitives, producing teleoperation-free trajectories with state, contact, and action data for base-VLA training and WM adaptation.

\subsection{Dense Progress Reward Models}
\label{sec:reward}
SARM~\cite{sarm} estimates progress hierarchically: it first predicts the current semantic stage, then conditions a second estimator on that stage to predict continuous within-stage completion. Dataset-level stage-duration priors combine both outputs into normalized task progress. SARM2~\cite{sarm2} generalizes this hierarchy across tasks with a mixture-of-experts stage estimator. Its rollout adaptation uses segment-level labels of \emph{fast progress}, \emph{slow progress}, \emph{adjust}, or \emph{mistake}, together with final task progress.

\subsection{DreamDojo: A Whole-Body-Control World Model}
\label{sec:dreamdojo}
DreamDojo~\cite{dreamdojo} is an action-conditioned video WM built on Cosmos~\cite{cosmos}. Given a single seed frame and a sequence of achieved 43-DoF joint states, it predicts future frames in fixed-length chunks,
$\hat{o}_{t+1:t+k}=W_\phi(o_t,\hat q_{t:t+k})$. Its post-training on real Unitree G1 WBC trajectories provides a whole-body prior, while autoregressive error accumulation~\cite{diffusionforcing,selfforcing,weaver} motivates broad play data and bounded rollout horizons.

\subsection{Reinforcement Learning for Flow-Matching Policies}
\label{sec:rl}
A flow-matching action head generates an action chunk by integrating a learned velocity field, but its implicit likelihood complicates standard policy gradients. We consider three alternatives. \textbf{DICE-RL}~\cite{sun2026prior} searches modes of a frozen flow prior $\pi_\text{pre}$ and learns a local residual correction:
\begin{equation}
  \begin{aligned}
    a_i &= \underbrace{\pi_\text{pre}(o,z_i)}_{\text{frozen prior}} + r_\psi(o,z_i),
    \quad z_i \sim \mathcal{N}(0,I), \\
    i^{\star} &= \arg\max_{1 \le i \le K} \; \min_{1 \le j \le M} Q_j(o,z_i,a_i),
  \end{aligned}
  \label{eq:residual}
\end{equation}
where $K$ candidate action chunks are evaluated by an $M$-member critic ensemble, and $a_{i^\star}$ is selected.
\textbf{AWR}~\cite{peng2019advantage} performs advantage-weighted supervised regression:
\begin{equation}
  \theta \leftarrow \arg\max_{\theta} \; \mathbb{E}_{(o,a)\sim\mathcal{D}}\Big[\log \pi_{\theta}(a\mid o)\,\exp\big(\tfrac{1}{\beta}A(o,a)\big)\Big],
  \label{eq:awr}
\end{equation}
where $A$ is the estimated advantage and $\beta$ is a temperature. \textbf{ReinFlow}~\cite{reinflow} converts the deterministic flow ODE into an SDE with learned noise; each discretized step is Gaussian:
\begin{equation}
  a^{(k+1)} \sim \mathcal{N}\big(a^{(k)} + v_\theta(a^{(k)},k)\,\Delta t, \; \sigma_\phi(a^{(k)},k,o)^2 I\big),
  \label{eq:reinflow}
\end{equation}
This gives tractable per-step log-likelihoods for on-policy optimization such as PPO~\cite{schulman2017proximal}.

\begin{figure*}[t!]
  \centering
  \includegraphics[width=\textwidth]{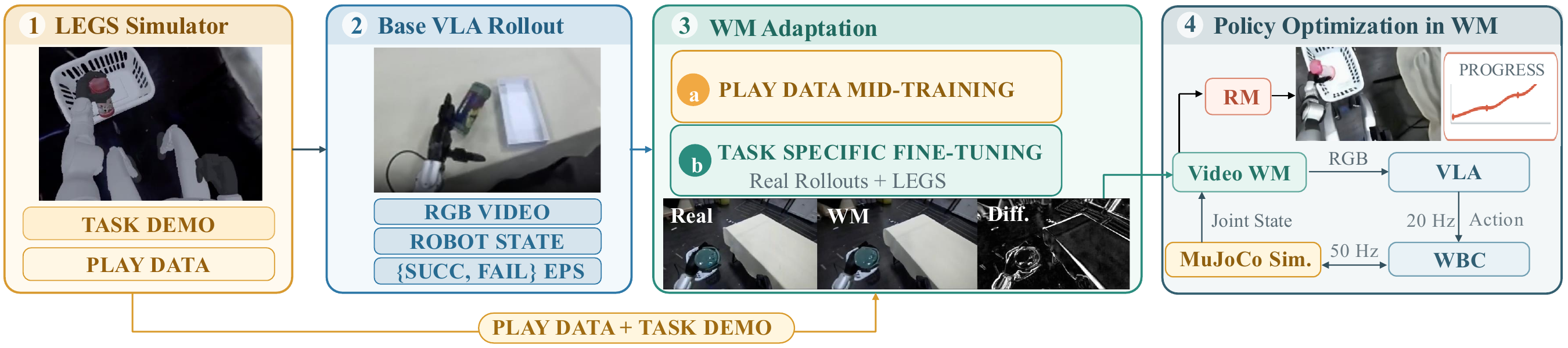}
  \caption{\textbf{\NAME{} pipeline.} The LEGS simulator supplies synthetic task demonstrations and play data. Autonomous real robot rollouts adapt the world model, which is coupled with the VLA and SONIC--MuJoCo WBC module for rollout collection and policy optimization under dense reward.}
  \label{fig:sys}
  \vspace{-0.35\baselineskip}
\end{figure*}

\section{Method}
\subsection{Real-to-Sim-to-Real Whole-Body VLA Initialization}
\label{sec:vla_init}
For each task, we construct a LEGS simulator and train a whole-body VLA with substantial hardware success, enabling teleoperation-free WM-adaptation rollouts.

\subsubsection{Task Construction and Randomization}
Using the robot-mounted ZED 2i camera~\cite{zed2i}, we reconstruct the deployment room and objects in the LEGS simulator and define task actions, success/failure criteria, and randomization. Object position is sampled within a $0.20\,\mathrm{m}\times0.20\,\mathrm{m}$ region with yaw in $[-180^\circ,180^\circ]$; the base starts within a $0.30\,\mathrm{m}\times0.30\,\mathrm{m}$ region centered $1.5$\,m from the table with yaw in $[-30^\circ,30^\circ]$. Camera roll and pitch are independently perturbed within $\pm15^\circ$.

\subsubsection{Base VLA Training}
We collect $200$ synthetic episodes per task in under $30$ minutes on 8 NVIDIA L40S GPUs, then fine-tune a task-specific GR00T N1.7 policy~\cite{gr00t} for $50{,}000$ steps, freezing the vision-language backbone. Training takes $3.5$ hours on 8 NVIDIA H100 GPUs.

\subsubsection{Autonomous Real Robot Rollouts}
\rv{The initial policy runs autonomously until we construct a balanced retained dataset of $20$ real robot trajectories per task, comprising $10$ successes and $10$ failures. Because the base policies have SR below $50\%$, collection can require more than $20$ attempts; excess failed trajectories are discarded and are not used for training. Each retained trajectory contains synchronized egocentric RGB, joint states, and VLA actions. The same retained dataset is used for both task-specific WM adaptation and the Real RL baseline, ensuring identical real robot training data. Balancing the outcomes prevents the WM adaptation set from being dominated by failures.}

\subsection{World Model Fine-Tuning}
\label{sec:wm}
We initialize from DreamDojo's released \texttt{2B\_G1\_post-train} checkpoint~\cite{dreamdojo} and train in two stages. \rv{LEGS simulator play data cover contact dynamics and locomotion-induced viewpoint changes; each task then uses $200$ synthetic demonstrations and the $20$ retained real robot trajectories described above to combine broad state-action coverage with deployment appearance.}

\subsubsection{Assets}
We use over $100$ household objects from open-source 3D object datasets~\cite{ycb, deitke2023objaverse, downs2022google, collins2022abo} and randomize their mass and friction across episodes.

\subsubsection{Manipulation Play Data}
Each two-minute manipulation episode samples 5--8 objects with randomized table poses and one active arm; the other remains idle. We perturb the nominal grasp target with independent uniform offsets of $\pm2$, $\pm3$, and $\pm2$\,cm and allow up to three grasp-and-place attempts per object. All outcomes, including misses, slips, collisions, and failed placements, are retained to supervise contact dynamics.

\subsubsection{Locomotion Play Data}
Each locomotion episode samples an approach to a $0.50$\,m standoff followed by a random $90^\circ$ turn, or an approach to $0.80$\,m followed by a clockwise or counterclockwise semicircle. Forward speed is $0.2$--$0.3$\,m/s, lateral velocity is zero, and SONIC tracks yaw. These trajectories expose the WM to large egocentric viewpoint changes caused by turning and sustained walking, which are sparse in task-specific rollouts.

\subsubsection{Play Data Mid-Training}
The corpus contains $400$ manipulation episodes ($10$ hours) and $100$ locomotion episodes ($1$ hour), totaling $500$ episodes and $11$ hours. Starting from \texttt{2B\_G1\_post-train}, we mid-train for $15{,}000$ steps and $10$ hours on 8 NVIDIA H100 GPUs, producing the shared checkpoint $\phi_{\mathrm{mid}}$.

\subsubsection{Task-Specific Fine-Tuning}
\rv{For each task $\tau$, we fine-tune $\phi_{\mathrm{mid}}$ on $200$ synthetic demonstrations and the same $20$ retained real robot trajectories. Synthetic trajectories cover randomized configurations and successful motions; real trajectories provide deployment-camera successes and failures that bridge the residual visual gap.} Fine-tuning runs for $30{,}000$ steps and $17$ hours per task, producing $\phi_\tau$ for policy rollout and optimization.

\subsection{Policy Rollout within the World Model}
\label{sec:rollout}
A rollout integrates the GR00T VLA, the SONIC--MuJoCo WBC, and the DreamDojo WM. These components exchange policy commands, achieved robot states, and generated visual observations at their respective execution rates, as illustrated in Fig.~\ref{fig:sys}.

\subsubsection{VLA Interface}
At policy step $t$, the flow-matching VLA receives an egocentric RGB observation $o_t \in \mathbb{R}^{H_i \times W_i \times 3}$, a proprioceptive state $s_t \in \mathbb{R}^{15}$, and a language instruction $\ell$. It predicts a chunk of $H=40$ whole-body targets at a $20$\,Hz execution rate:
\begin{equation}
  A_t = \big(a_t^1,\dots,a_t^{H}\big) \sim \pi_\theta\big(\cdot \mid o_t, s_t, \ell\big) \in \mathbb{R}^{H \times 18}, \quad H = 40,
  \label{eq:vlaio}
\end{equation}
We execute the first $H_e=24$ targets before querying the policy again, matching the execution horizon used on the real robot. Each target is represented in task space:
\begin{equation}
  a = \big(\underbrace{T_L,\gamma_L}_{7},\; \underbrace{T_R,\gamma_R}_{7},\; \underbrace{v_x,v_y,v_\psi,h}_{4}\big) \in \mathbb{R}^{18},
  \label{eq:action}
\end{equation}
where $T_L,T_R \in SE(3)$ denote hand poses represented by position and Euler angles, $\gamma_L,\gamma_R \in [0,1]$ denote gripper openings, $(v_x,v_y,v_\psi)$ are planar base-velocity commands, and $h$ is the humanoid height target. The proprioceptive input uses the corresponding command-space representation:
\begin{equation}
  s = \big(T_L,\gamma_L,\; T_R,\gamma_R,\; h\big) \in \mathbb{R}^{15},
  \label{eq:state}
\end{equation}
which contains the commanded hand poses, gripper openings, and torso height. Joint angles are not provided to the VLA.

\subsubsection{WBC State Evolution}
The SONIC--MuJoCo WBC maps each task-space target $a$ to achieved whole-body motion $q$. SONIC computes control outputs from the current joint state, and MuJoCo advances that state. Within each target interval, this feedback loop runs at $50$\,Hz:
\begin{equation}
  q_{k+1} = f\big(q_k,\, \pi_{\mathrm{wbc}}(q_k, a)\big), \quad q_k \in \mathbb{R}^{43}, \quad \Delta t_{\mathrm{wbc}} = \tfrac{1}{50}\,\mathrm{s},
  \label{eq:wbcstep}
\end{equation}
where $q_k$ is the $43$-dimensional joint state and $f$ denotes the MuJoCo transition. Within the WBC, MuJoCo closes the state-feedback loop required by SONIC and supplies the achieved joint state used to condition the WM. 

\subsubsection{World Model Observation Generation}
For each executed action chunk, the WM receives a single seed frame and a subsampled sequence of achieved joint states, and generates $C=12$ egocentric frames:
\begin{equation}
  \begin{gathered}
    \big(o^{(1)},\dots,o^{(C)}\big) = W_\phi\big(o^{(0)},\, \hat{q}^{(0:C)}\big), \\
    \hat{q}^{(m)} = \bar q_{\rho m}, \qquad C = 12,
  \end{gathered}
  \label{eq:wmgen}
\end{equation}
The WM is therefore conditioned on the achieved robot state rather than the requested target. Each generated frame is held for $\rho$ policy steps, such that $o_j=o^{(\lceil j/\rho\rceil)}$, and only the terminal generated frame seeds the next VLA query and WM generation call. \rv{This separation assigns robot-state evolution to the WBC and visual scene evolution, including object interactions, to the WM, which learns contact-rich transitions from simulated play data.}

Temporal alignment between the policy and WM is important because the WM is fine-tuned at $10$\,fps, whereas policy targets are executed at $20$\,Hz. We choose the stride $\rho$ so that the state displacement between consecutive WM frames matches the displacement observed during fine-tuning:
\begin{equation}
  \begin{aligned}
    \mathbb{E}\big\| \hat{q}^{(m+1)} - \hat{q}^{(m)} \big\|
      &= \rho\, \mathbb{E}\big\| \bar q_{j+1} - \bar q_j \big\| \\
      &\approx \mathbb{E}_{\mathcal{D}_{\mathrm{ft}}}\big\| \Delta q \big\|,
  \end{aligned}
  \label{eq:stride}
\end{equation}
which yields $\rho=2$. Without this subsampling, consecutive conditioning states contain less motion than the WM fine-tuning data and produce nearly static predictions. Finally, each rollout is initialized from a real observation, $o^{(0)} \sim \mathcal{D}_{\mathrm{real}}$, anchoring generation to the deployment-domain appearance.

\subsection{Reward Estimation on Generated Video}
\label{sec:rm}
\subsubsection{Reward Model Training}
We train one task-specific SARM model per task, dividing each trajectory into locomotion and manipulation stages and supervising piecewise-linear progress between fixed milestones. Synthetic stage boundaries come from executed primitives; failed real rollouts follow SARM2, with progress rising to the reached stage and dropping to zero at the annotated failure. The predictor is an $8$-layer, $12$-head transformer ($d_{\mathrm{model}}=768$) over frozen DINOv3~\cite{simeoni2025dinov3} visual features. It conditions on the task instruction and a $16$-frame history, with up to $8$ rewind steps enforcing decreasing progress under backward playback. Training uses the WM rollout cadence of $10$\,fps.

\subsubsection{Dense Rewards and Termination}
\begin{figure}[t!]
  \centering
  \includegraphics[width=\columnwidth]{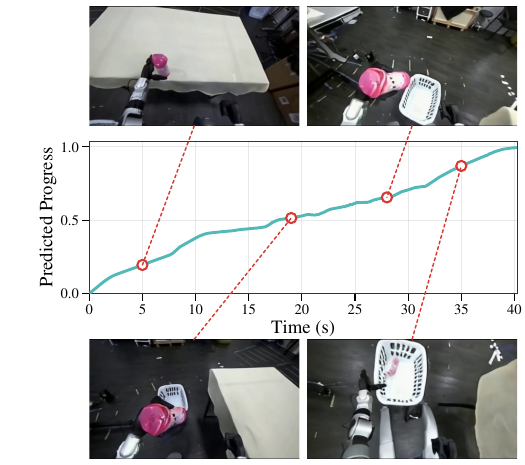}
  \caption{\textbf{Dense progress on a world model rollout.} Red circles link four \textbf{WM-generated} views to their progress estimates. Increments of the smoothed progress prediction provide dense rewards.}
  \label{fig:wm_progress}
  \vspace{-0.35\baselineskip}
\end{figure}

For a generated rollout, the model predicts progress $p_t$. We smooth this estimate with an exponential moving average and define the per-step reward as its increment:
\begin{equation}
  r_t = \tilde{p}_t - \tilde{p}_{t-1}, \qquad \tilde{p}_t = \alpha \tilde{p}_{t-1} + (1-\alpha) p_t,
  \label{eq:reward}
\end{equation}
where $\alpha=0.7$ suppresses frame-level prediction noise. Generated rollouts extend to $1.5\times$ the real robot episode length to accommodate slower policies. We terminate a rollout and mark it successful when $\tilde{p}_t \ge 0.9$ for 5 consecutive inferences. Frames after this point are discarded, preventing post-completion generation from contributing an idle tail to policy training.
Figure~\ref{fig:wm_progress} visualizes this signal over a representative WM rollout, with sampled frames linked to their corresponding progress estimates.

\subsection{Reinforcement-Learning Instantiations}
We instantiate DICE-RL, ReinFlow, and AWR as policy-improvement algorithms within the same WM rollout and reward interface. \rv{All use the same progress predictions with algorithm-specific reward transformations.} In each round, the current VLA supplies actions for WM rollouts, and the selected algorithm updates its action head from trajectories collected under the current policy. The vision-language backbone remains frozen, with its pooled features cached for action-head and value-function training.

\subsubsection{DICE-RL}
We freeze the base VLA and train its residual actor with a $20$-member critic ensemble. To stabilize noisy dense rewards, we augment the temporal-difference target with a Monte Carlo return from the sparse rollout outcome:
\begin{equation}
  y_t = \rv{r_t^{\mathrm{D}}} + \gamma\min_j Q_j^-
  + wR_t^{\mathrm{mc}}, \qquad w=0.5,
  \label{eq:critictarget}
\end{equation}
\rv{DICE-RL uses the nonnegative critic reward $r_t^{\mathrm{D}}=\max(0,\tilde p_t-\tilde p_{t-1})$ to reduce sensitivity to local progress noise; $Q_j^-$ is target critic $j$ at the next state--action pair.} The Monte Carlo term is $R_t^{\mathrm{mc}}=\sum_{k=t}^{T-1}\gamma^{k-t}r_k^{\mathrm{sp}}$, where $r_k^{\mathrm{sp}}\in\{0,1\}$ is one only at the final step of a successful rollout and zero otherwise. We regularize the residual and evaluate $K=8$ noise-conditioned chunks per actor update; $\gamma=0.995$, the batch size is $128$, and both learning rates are $10^{-4}$.

\subsubsection{ReinFlow}
Unlike DICE-RL, ReinFlow updates the flow policy itself. We fully fine-tune the VLA's flow-matching action head, while keeping the vision-language backbone frozen. The Gaussian transitions in Eq.~(\ref{eq:reinflow}) provide tractable per-step likelihoods for a clipped PPO objective. We additionally anchor the fine-tuned action to the frozen base policy and penalize changes in its temporal differences:
\begin{equation}
  \begin{aligned}
    \mathcal{L}_{\mathrm{RF}}(\theta)
    &= -\mathbb{E}\!\left[
      \min\!\left(\rho\hat{A},
      \operatorname{clip}(\rho,1-\varepsilon,1+\varepsilon)\hat{A}\right)
    \right] \\
    &\quad 
    + \beta_{\mathrm{BC}}\mathbb{E}\!\left[
      \lVert a_\theta-\bar{a}\rVert_2^2
      + \lVert\delta a_\theta-\delta\bar{a}\rVert_2^2
    \right],
  \end{aligned}
  \label{eq:rfobj}
\end{equation}
where $\rho=\pi_\theta(a\mid o)/\pi_{\theta_{\mathrm{old}}}(a\mid o)$, $\hat{A}$ is estimated by the state-value critic, $\bar{a}$ is the frozen-policy action, and $\delta a_t=a_{t+1}-a_t$. 

\subsubsection{AWR}
We use the advantage-conditioned AWR formulation of $\pi^{*}_{0.6}$~\cite{intelligence2025pi}, which replaces likelihood-weighted regression with conditioning on a binary improvement variable:
\begin{equation}
  \begin{aligned}
    p\!\left(I\mid A^{\pi_{\mathrm{ref}}}(o,a,\ell)\right)
    &= \delta\!\left(A^{\pi_{\mathrm{ref}}}(o,a,\ell)>\epsilon_\ell\right), \\
    \hat{\pi}(a\mid o,\ell)
    &\propto \pi_{\mathrm{ref}}(a\mid o,\ell)
    \left[
      \frac{\pi_{\mathrm{ref}}(a\mid I,o,\ell)}
           {\pi_{\mathrm{ref}}(a\mid o,\ell)}
    \right]^\beta .
  \end{aligned}
  \label{eq:advcond}
\end{equation}
We estimate $A^{\pi_{\mathrm{ref}}}=r_t+\gamma V(o_{t+1})-V(o_t)$, set $\epsilon_\ell=0$ and $\beta=1$, and append the resulting label to the language input. The policy is trained on both positive and negative chunks with the original flow-matching objective and conditioned on improvement at deployment.

\subsection{Closed-Loop Policy Improvement}
\label{sec:loop}
\subsubsection{Implementation Details}
After task-specific fine-tuning, we freeze the WM and optimize each task for $15$ rounds. Each round generates $32$ current-policy episodes from real seed frames for up to $60$\,s. The progress model labels each video and truncates successful episodes at completion (Sec.~\ref{sec:rm}), after which we perform $10^3$ updates. Policy, critic, target-network, and optimizer states persist across rounds while the rollout batch is refreshed, yielding $480$ episodes and $1.5\times10^4$ updates per task. We evaluate every checkpoint on the real robot. Each run uses 8 NVIDIA H100 GPUs with eight parallel workers, each hosting an independent WM, WBC, and MuJoCo instance. A full run takes about $12$ hours: $75\%$ for video generation, $20\%$ for RL updates, and the remainder for progress labeling and feature caching.

\section{Experiments}
We ask three research questions in this section: 
\vspace{-0.2\baselineskip}
\begin{itemize}
\setlength{\topsep}{0.2\baselineskip}
\setlength{\itemsep}{0pt}
\setlength{\parsep}{0pt}
\item \textbf{Q1:} How do simulated whole-body play data affect WM fidelity and downstream policy learning? 
\item \textbf{Q2:} \rq{Why is a video WM a better policy-improvement environment than the simulator?}
\item \textbf{Q3:} \rq{How should policies learn effectively from WM rollouts?}
\end{itemize}
\vspace{-0.35\baselineskip}

\begin{figure*}[h!]
  \centering
  \includegraphics[width=\textwidth]{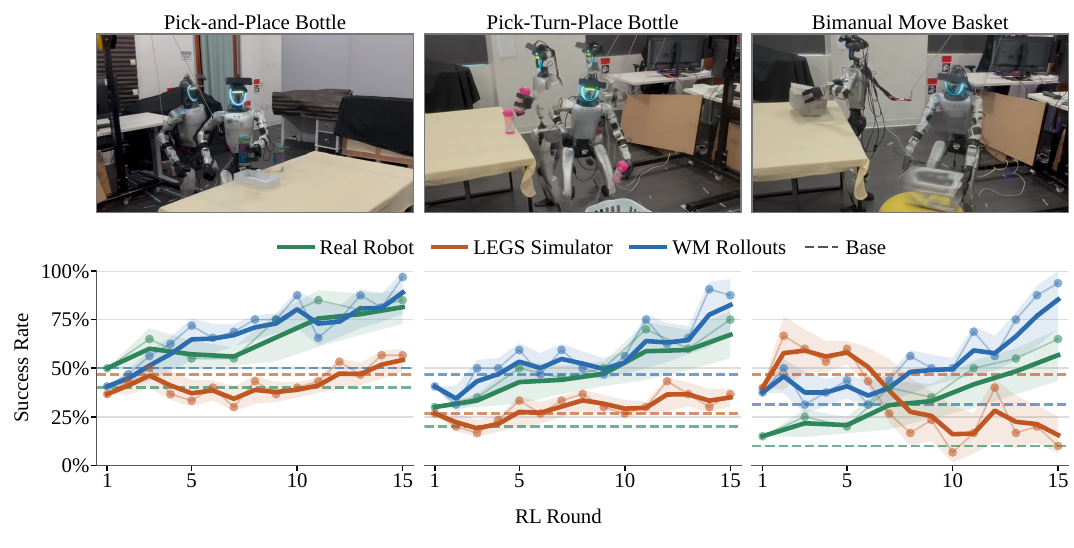}
  \caption{\textbf{DICE-RL improvement across three loco-manipulation tasks.} Top: representative real robot executions. Bottom: success over $15$ RL rounds on the real robot, in the LEGS simulator, and in world model rollouts. Solid lines are exponential moving averages, faint lines and markers are per-round measurements, bands are exponentially weighted standard deviations, and dashed lines show base-BC success. \rv{WM success is estimated by the progress model, and the bands show variation across RL rounds rather than uncertainty across independent runs.}}
  \label{fig:online_rl}
  \vspace{-0.35\baselineskip}
\end{figure*}

\subsection{Experimental Setup}
\subsubsection{Tasks}
We evaluate three real robot tasks with increasing locomotion--manipulation coordination (Fig.~\ref{fig:online_rl}). In \emph{Pick-and-Place Bottle} (PNP), the robot walks to a table, grasps a bottle with its left hand, and places it upright in a box. \emph{Pick-Turn-Place Bottle} (PTP) adds post-grasp locomotion: the robot steps back, turns about $90^\circ$, approaches a floor basket, squats, and places the bottle inside. In \emph{Bimanual Move Basket} (Basket), it approaches a side platform, squats to lift a basket bimanually, returns to the table, and places it. The tasks span stationary manipulation, object transport, and bimanual control.

\subsubsection{World Model Evaluation}
For each task, we compare two WMs under a matched budget of $45{,}000$ optimization steps: the full model receives $15{,}000$ steps of mid-training on play data followed by $30{,}000$ task-specific steps, while the ablation skips play data and uses all $45{,}000$ steps for task-specific fine-tuning. \rv{Both use the same $200$ synthetic episodes and $20$ retained real robot trajectories. A separate held-out set contains $20$ synthetic and $20$ real episodes per task, each split evenly between success and failure, and is used only to report WM fidelity.} We replay each recorded action trajectory and compare the generated frames with the temporally aligned ground truth, excluding terminal-frame padding. We report PSNR, interaction PSNR (iPSNR), SSIM, LPIPS, and FID. iPSNR is computed over the union of robot-hand and manipulated-object masks obtained with Grounding DINO~\cite{groundingdino} and SAM 2~\cite{sam2}. The same frames and masks are used for both models; higher PSNR, iPSNR, and SSIM are better, while lower LPIPS and FID are better.

\subsubsection{Policy Evaluation}
For each policy-task pair, we report success rate (SR) over $20$ real robot trials. \rv{Table~\ref{tab:policy_results} reports the final checkpoint from one training run per method and task, evaluated over these $20$ trials without checkpoint selection on hardware results.} Success requires the bottle to stand upright in the tabletop box for Pick-and-Place Bottle, the bottle to be placed in the floor basket for Pick-Turn-Place Bottle, or the bimanually carried basket to be placed stably on the table for Bimanual Move Basket. To track learning progress, we also evaluate intermediate policies on the real robot for every two rounds and in the LEGS simulator and WM rollouts (Fig.~\ref{fig:online_rl}) every round. Success in the LEGS simulator is determined by full-state task predicates, whereas success in WM rollouts is determined by the progress reward model following the protocol mentioned in Sec.~\ref{sec:rm}.

\subsubsection{Baselines and Ablations}
All methods start from the same synthetic-data BC checkpoint. \textbf{BC} receives no RL post-training. \rv{\textbf{Real RL} applies DICE-RL to the same $20$ retained real robot trajectories used for task-specific WM fine-tuning.} \textbf{Sim RL} applies DICE-RL directly in the LEGS simulator. Among WM-based baselines, \textbf{VLAW}~\cite{vlaw} behavior-clones successful WM rollouts, whereas \textbf{WMPO}~\cite{wmpo} uses GRPO with sparse terminal rewards. Because our VLA has a flow-matching action head, we implement WMPO with Flow-GRPO~\cite{flowgrpo}. We instantiate our pipeline with AWR, ReinFlow, and DICE-RL. All ablations use DICE-RL, the best-performing of these three RL algorithms. \textbf{Sparse Reward} replaces dense progress with terminal success feedback. \textbf{Offline} uses a fixed set of $480$ WM rollouts ($15$ rounds $\times$ $32$ episodes), generated once by the base BC policy, matching the total rollout budget without refreshing data as the policy changes. \textbf{Without play data} omits WM mid-training on play data.

\subsection{Q1: Effect of WM Mid-Training on Play Data}
Table~\ref{tab:wm_fidelity} shows a consistent benefit from play data: the play-data model outperforms the base initialization on every metric across all three tasks. Averaged across tasks, global PSNR improves by $0.93$\,dB and interaction-region iPSNR by $1.17$\,dB, with the iPSNR gain larger on every task. Because iPSNR focuses on the robot hands and manipulated object, this larger improvement suggests that play data particularly strengthens the modeling of contact-rich physical interactions and object dynamics, rather than merely improving global scene appearance.

The policy ablation in Table~\ref{tab:policy_results} shows a corresponding downstream effect. Removing play data from the DICE-RL pipeline reduces SR on all three tasks. The mean success rate falls from $75.0\%$ to $63.3\%$. Play data mid-training therefore improves both WM fidelity and downstream policy performance.

\begin{table}[t]
  \centering
  \caption{\textbf{World model fidelity with and without play data.} Base uses $45{,}000$ task-specific steps; Play uses $15{,}000$ play data mid-training steps followed by $30{,}000$ task-specific steps. Both use identical task data and held-out episodes; best values are bold.}
  \label{tab:wm_fidelity}
  \footnotesize
  \setlength{\tabcolsep}{2.3pt}
  \renewcommand{\arraystretch}{1.03}
  \begin{tabular}{lcccccc}
    \toprule
    Task & Variant & PSNR $\uparrow$ & iPSNR $\uparrow$ & SSIM $\uparrow$ & LPIPS $\downarrow$ & FID $\downarrow$ \\
    \midrule
    \multirow{2}{*}{PNP}
      & Base & 20.22 & 15.12 & 0.75 & 0.24 & 32.61 \\
      & Play & \textbf{21.46} & \textbf{16.71} & \textbf{0.77} & \textbf{0.23} & \textbf{30.97} \\
    \midrule
    \multirow{2}{*}{PTP}
      & Base & 18.38 & 14.96 & 0.70 & 0.29 & 33.08 \\
      & Play & \textbf{19.08} & \textbf{15.86} & \textbf{0.71} & \textbf{0.28} & \textbf{31.97} \\
    \midrule
    \multirow{2}{*}{Basket}
      & Base & 17.70 & 13.06 & 0.67 & 0.30 & 37.54 \\
      & Play & \textbf{18.55} & \textbf{14.07} & \textbf{0.68} & \textbf{0.29} & \textbf{37.42} \\
    \bottomrule
  \end{tabular}
\end{table}

\subsection{Q2: WM for Policy Learning}
\rq{Table~\ref{tab:policy_results} compares policy improvement on hardware, in the LEGS simulator, and in the video WM.} \rv{AWR, ReinFlow, and DICE-RL all achieve higher mean real robot SR than BC, Sim RL, and Real RL. DICE-RL performs best overall, reaching a mean SR of $75\%$ across the three tasks. From the same real robot training data used by Real RL, the video WM supports $480$ iteratively collected online policy rollouts without further hardware interaction, demonstrating the value of scalable WM experience.}

Despite using the same LEGS simulator as BC, Sim RL harms SR on all three tasks. \rv{This degradation is consistent with RL exploiting the residual visual sim-to-real gap in the LEGS simulator.} The learning curves in Fig.~\ref{fig:online_rl} provide complementary evidence: WM and real robot success rates track one another and rise over optimization, whereas success in the LEGS simulator remains nearly level on PNP and PTP, and decreases substantially on Basket.

\rq{Together, these results clarify the environments' complementary roles. The LEGS simulator provides structured demonstrations and physically grounded coverage, but the residual visual sim-to-real gap makes it unreliable for final policy optimization. By incorporating real robot rollouts, the video WM generates observations and trajectories that better match those encountered on hardware, making it a more useful environment for policy improvement.}

\begin{table}[t]
  \centering
  \caption{\textbf{Real robot policy SR over $20$ trials for 3 tasks.} Ablations use DICE-RL; best results are bold.}
  \label{tab:policy_results}
  \footnotesize
  \setlength{\tabcolsep}{2pt}
  \renewcommand{\arraystretch}{1.0}
  \begin{tabular}{@{}c@{\hspace{5pt}}c@{}}
    \begin{tabular}[t]{@{}lccc@{}}
      \toprule
      \multicolumn{4}{c}{Baselines} \tabularnewline
      \midrule
      Method & PNP & PTP & Basket \tabularnewline
      \midrule
      BC      & 40\% & 20\% & 10\% \tabularnewline
      Real RL & 60\% & 45\% & 40\% \tabularnewline
      Sim RL  & 30\% & 0\%  & 0\% \tabularnewline
      VLAW    & 45\% & 20\% & 35\% \tabularnewline
      WMPO    & 55\% & 30\% & 35\% \tabularnewline
      \bottomrule
    \end{tabular}
    &
    \begin{tabular}[t]{@{}lccc@{}}
      \toprule
      \multicolumn{4}{c}{RL Algorithms} \tabularnewline
      \midrule
      Method   & PNP & PTP & Basket \tabularnewline
      \midrule
      AWR      & 75\% & 70\% & \textbf{70\%} \tabularnewline
      ReinFlow & 60\% & 50\% & 45\% \tabularnewline
      DICE-RL  & \textbf{85\%} & \textbf{75\%} & 65\% \tabularnewline
      \midrule
      \multicolumn{4}{c}{Ablations} \tabularnewline
      \midrule
      Sparse   & 65\% & 40\% & 30\% \tabularnewline
      Offline  & 70\% & 55\% & 50\% \tabularnewline
      w/o Play & 70\% & 65\% & 55\% \tabularnewline
      \bottomrule
    \end{tabular}
    \tabularnewline
  \end{tabular}
\end{table}

\subsection{Q3: How Should Policies Learn from WM Rollouts?}
Effective policy improvement requires learning from the full rollout distribution. VLAW behavior-clones only successful WM rollouts and achieves an average SR of $33.3\%$, whereas DICE-RL learns from successful, failed, and suboptimal episodes and reaches $75.0\%$. Dense progress assigns graded credit before task completion, allowing each rollout to contribute to learning.

Dense reward is particularly important because WM rollout generation is computationally intensive. WMPO uses sparse terminal rewards and achieves an average SR of $40.0\%$, utilizing only terminal outcome label from each trajectory. The original paper's ablation~\cite{wmpo} further shows substantially larger gains with $32$ H100 GPUs than with $8$, the budget used here. Consistently, replacing dense progress with sparse feedback in DICE-RL lowers mean success from $75.0\%$ to $45.0\%$.

The rollout buffer must track the evolving policy distribution. In the Offline ablation, a static buffer of $480$ trajectories is generated once by the base behavior policy $\pi_{\mathrm{BC}}$ and reused under the same training-step budget. As the actor departs from $\pi_{\mathrm{BC}}$, the growing policy--data distribution shift reduces mean SR to $58.3\%$, compared with $75.0\%$ when trajectories are recollected under the current policy each round. The gains are robust across policy-update rules: AWR and DICE-RL achieve mean SRs of $71.7\%$ and $75.0\%$, while ReinFlow exhibits less stable optimization but still exceeds both BC and Real RL in mean SR. Together, these results identify full-distribution learning, dense rewards, and iterative online rollout collection as key ingredients for effective WM-based RL.

\section{Conclusion}
\rq{We presented \NAME{}, a teleoperation-free framework for whole-body humanoid VLA post-training. Its central design assigns complementary roles to two environments: the simulator supplies policy initialization and broad, physically grounded experience, while the video WM incorporates limited real robot deployment experience and supports policy improvement. This separation matters empirically: direct simulator RL reduces success on all three tasks, whereas video WM RL raises mean SR from $23.3\%$ to $75.0\%$, compared with $48.3\%$ for real robot RL. Dense progress rewards and online rollout collection further enable effective learning from generated trajectories.}

Limitations include task-specific adaptation. Unitree G1-specific interfaces preclude zero-shot transfer to other humanoids. Future work should address cross-embodiment transfer, long-horizon multi-view prediction, rollout efficiency, and joint policy-WM adaptation.

\bibliographystyle{IEEEtran}
\bibliography{ref}

@article{openvla,
  title={Openvla: An open-source vision-language-action model},
  author={Kim, Moo Jin and Pertsch, Karl and Karamcheti, Siddharth and Xiao, Ted and Balakrishna, Ashwin and Nair, Suraj and Rafailov, Rafael and Foster, Ethan and Lam, Grace and Sanketi, Pannag and others},
  journal={arXiv preprint arXiv:2406.09246},
  year={2024}
}

@article{rt2,
  title={Rt-2: Vision-language-action models transfer web knowledge to robotic control},
  author={Brohan, Anthony and Brown, Noah and Carbajal, Justice and Chebotar, Yevgen and Chen, Xi and Choromanski, Krzysztof and Ding, Tianli and Driess, Danny and Dubey, Avinava and Finn, Chelsea and others},
  journal={arXiv preprint arXiv:2307.15818},
  year={2023}
}

@article{pi0,
  title={{$\pi_0$}: A Vision-Language-Action Flow Model for General Robot Control},
  author={Black, Kevin and Brown, Noah and Driess, Danny and Esmail, Adnan and Equi, Michael and Finn, Chelsea and Fusai, Niccolo and Groom, Lachy and Hausman, Karol and Ichter, Brian and others},
  journal={arXiv preprint arXiv:2410.24164},
  year={2024}
}

@article{pi05,
  title={{$\pi_{0.5}$}: A Vision-Language-Action Model with Open-World Generalization},
  author={Intelligence, Physical and Black, Kevin and Brown, Noah and Darpinian, James and Dhabalia, Karan and Driess, Danny and Esmail, Adnan and Equi, Michael and Finn, Chelsea and Fusai, Niccolo and others},
  journal={arXiv preprint arXiv:2504.16054},
  year={2025}
}

@article{gr00t,
  title={Gr00t n1: An open foundation model for generalist humanoid robots},
  author={Bjorck, Johan and Casta{\~n}eda, Fernando and Cherniadev, Nikita and Da, Xingye and Ding, Runyu and Fan, Linxi and Fang, Yu and Fox, Dieter and Hu, Fengyuan and Huang, Spencer and others},
  journal={arXiv preprint arXiv:2503.14734},
  year={2025}
}

@article{deitke2023objaverse,
  title={Objaverse-xl: A universe of 10m+ 3d objects},
  author={Deitke, Matt and Liu, Ruoshi and Wallingford, Matthew and Ngo, Huong and Michel, Oscar and Kusupati, Aditya and Fan, Alan and Laforte, Christian and Voleti, Vikram and Gadre, Samir Yitzhak and others},
  journal={Advances in Neural Information Processing Systems},
  volume={36},
  pages={35799--35813},
  year={2023}
}

@inproceedings{downs2022google,
  title={Google scanned objects: A high-quality dataset of 3d scanned household items},
  author={Downs, Laura and Francis, Anthony and Koenig, Nate and Kinman, Brandon and Hickman, Ryan and Reymann, Krista and McHugh, Thomas B and Vanhoucke, Vincent},
  booktitle={2022 international conference on robotics and automation (ICRA)},
  pages={2553--2560},
  year={2022},
  organization={Ieee}
}

@inproceedings{collins2022abo,
  title={Abo: Dataset and benchmarks for real-world 3d object understanding},
  author={Collins, Jasmine and Goel, Shubham and Deng, Kenan and Luthra, Achleshwar and Xu, Leon and Gundogdu, Erhan and Zhang, Xi and Vicente, Tomas F Yago and Dideriksen, Thomas and Arora, Himanshu and others},
  booktitle={2022 IEEE/CVF Conference on Computer Vision and Pattern Recognition (CVPR)},
  pages={21094--21104},
  year={2022},
  organization={IEEE}
}

@article{hu2023imitation,
  title={Imitation bootstrapped reinforcement learning},
  author={Hu, Hengyuan and Mirchandani, Suvir and Sadigh, Dorsa},
  journal={arXiv preprint arXiv:2311.02198},
  year={2023}
}

@article{huang2024mentor,
  title={Mentor: Mixture-of-experts network with task-oriented perturbation for visual reinforcement learning},
  author={Huang, Suning and Zhang, Zheyu and Liang, Tianhai and Xu, Yihan and Kou, Zhehao and Lu, Chenhao and Xu, Guowei and Xue, Zhengrong and Xu, Huazhe},
  journal={arXiv preprint arXiv:2410.14972},
  year={2024}
}

@inproceedings{yang2024robot,
  title={Robot fine-tuning made easy: Pre-training rewards and policies for autonomous real-world reinforcement learning},
  author={Yang, Jingyun and Mark, Max Sobol and Vu, Brandon and Sharma, Archit and Bohg, Jeannette and Finn, Chelsea},
  booktitle={2024 IEEE International Conference on Robotics and Automation (ICRA)},
  pages={4804--4811},
  year={2024},
  organization={IEEE}
}

@article{chen2025conrft,
  title={Conrft: A reinforced fine-tuning method for vla models via consistency policy},
  author={Chen, Yuhui and Tian, Shuai and Liu, Shugao and Zhou, Yingting and Li, Haoran and Zhao, Dongbin},
  journal={arXiv preprint arXiv:2502.05450},
  year={2025}
}

@article{wu2025robocopilot,
  title={Robocopilot: Human-in-the-loop interactive imitation learning for robot manipulation},
  author={Wu, Philipp and Shentu, Yide and Liao, Qiayuan and Jin, Ding and Guo, Menglong and Sreenath, Koushil and Lin, Xingyu and Abbeel, Pieter},
  journal={arXiv preprint arXiv:2503.07771},
  year={2025}
}

@article{intelligence2025pi,
  title     = {{$\pi^{*}_{0.6}$}: A VLA That Learns From Experience},
  author    = {{Physical Intelligence} and Amin, Ali and Aniceto, Raichelle and Balakrishna, Ashwin and Black, Kevin and Conley, Ken and Connors, Grace and Darpinian, James and Dhabalia, Karan and others},
  journal   = {arXiv preprint arXiv:2511.14759},
  year      = {2025}
}

@article{peng2019advantage,
  title={Advantage-weighted regression: Simple and scalable off-policy reinforcement learning},
  author={Peng, Xue Bin and Kumar, Aviral and Zhang, Grace and Levine, Sergey},
  journal={arXiv preprint arXiv:1910.00177},
  year={2019}
}

@article{lei2025rl,
  title={Rl-100: Performant robotic manipulation with real-world reinforcement learning},
  author={Lei, Kun and Li, Huanyu and Yu, Dongjie and Wei, Zhenyu and Guo, Lingxiao and Jiang, Zhennan and Wang, Ziyu and Liang, Shiyu and Xu, Huazhe},
  journal={arXiv preprint arXiv:2510.14830},
  year={2025}
}

@article{schulman2017proximal,
  title={Proximal policy optimization algorithms},
  author={Schulman, John and Wolski, Filip and Dhariwal, Prafulla and Radford, Alec and Klimov, Oleg},
  journal={arXiv preprint arXiv:1707.06347},
  year={2017}
}

@article{wagenmaker2025steering,
  title={Steering your diffusion policy with latent space reinforcement learning},
  author={Wagenmaker, Andrew and Nakamoto, Mitsuhiko and Zhang, Yunchu and Park, Seohong and Yagoub, Waleed and Nagabandi, Anusha and Gupta, Abhishek and Levine, Sergey},
  journal={arXiv preprint arXiv:2506.15799},
  year={2025}
}

@article{xiao2025self,
  title={Self-improving vision-language-action models with data generation via residual rl},
  author={Xiao, Wenli and Lin, Haotian and Peng, Andy and Xue, Haoru and He, Tairan and Xie, Yuqi and Hu, Fengyuan and Wu, Jimmy and Luo, Zhengyi and Fan, Linxi and others},
  journal={arXiv preprint arXiv:2511.00091},
  year={2025}
}

@article{ankile2025residual,
  title={Residual off-policy rl for finetuning behavior cloning policies},
  author={Ankile, Lars and Jiang, Zhenyu and Duan, Rocky and Shi, Guanya and Abbeel, Pieter and Nagabandi, Anusha},
  journal={arXiv preprint arXiv:2509.19301},
  year={2025}
}

@article{sun2026prior,
  title={From Prior to Pro: Efficient Skill Mastery via Distribution Contractive RL Finetuning},
  author={Sun, Zhanyi and Song, Shuran},
  journal={arXiv preprint arXiv:2603.10263},
  year={2026}
}

@article{reinflow,
  title={Reinflow: Fine-tuning flow matching policy with online reinforcement learning},
  author={Zhang, Tonghe and Yu, Chao and Su, Sichang and Wang, Yu},
  journal={Advances in Neural Information Processing Systems},
  volume={38},
  pages={106282--106319},
  year={2026}
}

@article{flowgrpo,
  title={Flow-{GRPO}: Training Flow Matching Models via Online Reinforcement Learning},
  author={Liu, Jie and Liu, Gongye and Liang, Jiajun and Li, Yangguang and Liu, Jiaheng and Wang, Xintao and Wan, Pengfei and Zhang, Di and Ouyang, Wanli},
  journal={arXiv preprint arXiv:2505.05470},
  year={2025}
}

@article{cosmos,
  title={Cosmos world foundation model platform for physical ai},
  author={Agarwal, Niket and Ali, Arslan and Bala, Maciej and Balaji, Yogesh and Barker, Erik and Cai, Tiffany and Chattopadhyay, Prithvijit and Chen, Yongxin and Cui, Yin and Ding, Yifan and others},
  journal={arXiv preprint arXiv:2501.03575},
  year={2025}
}

@article{dreamdojo,
  title   = {DreamDojo: A Generalist Robot World Model from Large-Scale Human Videos},
  author  = {Gao, Shenyuan and Liang, William and Zheng, Kaiyuan and Malik, Ayaan and Ye, Seonghyeon and Yu, Sihyun and Tseng, Wei-Cheng and Dong, Yuzhu and Mo, Kaichun and Lin, Chen-Hsuan and others},
  journal = {arXiv preprint arXiv:2602.06949},
  year    = {2026}
}

@inproceedings{wu2022daydreamer,
  title     = {DayDreamer: World Models for Physical Robot Learning},
  author    = {Wu, Philipp and Escontrela, Alejandro and Hafner, Danijar and Goldberg, Ken and Abbeel, Pieter},
  booktitle = {Conference on Robot Learning (CoRL)},
  year      = {2022}
}

@inproceedings{dreamer,
  title={Dream to Control: Learning Behaviors by Latent Imagination},
  author={Hafner, Danijar and Lillicrap, Timothy and Ba, Jimmy and Norouzi, Mohammad},
  booktitle={International Conference on Learning Representations (ICLR)},
  year={2020}
}

@inproceedings{drema,
  title={Dream to manipulate: Compositional world models empowering robot imitation learning with imagination},
  author={Barcellona, Leonardo and Zadaianchuk, Andrii and Allegro, Davide and Papa, Samuele and Ghidoni, Stefano and Gavves, Efstratios},
  booktitle={International Conference on Learning Representations},
  volume={2025},
  pages={56729--56763},
  year={2025}
}

@inproceedings{wmpo,
  title={Wmpo: World model-based policy optimization for vision-language-action models},
  author={Zhu, Fangqi and Yan, Zhengyang and Hong, Zicong and Shou, Quanxin and Ma, Xiao and Guo, Song},
  booktitle={International Conference on Learning Representations},
  volume={2026},
  pages={62486--62502},
  year={2026}
}

@article{vlaw,
  title={Vlaw: Iterative co-improvement of vision-language-action policy and world model},
  author={Guo, Yanjiang and Lee, Tony and Shi, Lucy Xiaoyang and Chen, Jianyu and Liang, Percy and Finn, Chelsea},
  journal={arXiv preprint arXiv:2602.12063},
  year={2026}
}

@article{wovr,
  title={Wovr: World models as reliable simulators for post-training vla policies with rl},
  author={Jiang, Zhennan and Zhou, Shangqing and Jiang, Yutong and Huang, Zefang and Wei, Mingjie and Chen, Yuhui and Zhou, Tianxing and Guo, Zhen and Lin, Hao and Zhang, Quanlu and others},
  journal={arXiv preprint arXiv:2602.13977},
  year={2026}
}

@article{worldgymnast,
  title={World-gymnast: Training robots with reinforcement learning in a world model},
  author={Sharma, Ansh Kumar and Sun, Yixiang and Lu, Ninghao and Zhang, Yunzhe and Liu, Jiarao and Yang, Sherry},
  journal={arXiv preprint arXiv:2602.02454},
  year={2026}
}

@article{rise,
  title={Rise: Self-improving robot policy with compositional world model},
  author={Yang, Jiazhi and Lin, Kunyang and Li, Jinwei and Zhang, Wencong and Lin, Tianwei and Wu, Longyan and Su, Zhizhong and Zhao, Hao and Zhang, Ya-Qin and Chen, Li and others},
  journal={arXiv preprint arXiv:2602.11075},
  year={2026}
}

@article{worldsample,
  title={WorldSample: Closed-loop Real-robot RL with World Modelling},
  author={Xue, Yuquan and Xu, Le and Liu, Zeyi and Wu, Zhenyu and Gu, Zhengyi and Song, Xinyang and Jia, Bofang and Wang, Ziwei},
  journal={arXiv preprint arXiv:2607.02431},
  year={2026}
}

@article{world4rl,
  title={World4rl: Diffusion world models for policy refinement with reinforcement learning for robotic manipulation},
  author={Jiang, Zhennan and Liu, Kai and Qin, Yuxin and Tian, Shuai and Zheng, Yupeng and Zhou, Mingcai and Yu, Chao and Li, Haoran and Zhao, Dongbin},
  journal={arXiv preprint arXiv:2509.19080},
  year={2025}
}

@article{playworld,
  title   = {PlayWorld: Learning Robot World Models from Autonomous Play},
  author  = {Yin, Tenny and Mei, Zhiting and Zheng, Zhonghe and Yamane, Miyu and Wang, David and Sceats, Jade and Bateman, Samuel M. and Zha, Lihan and Badithela, Apurva and Shorinwa, Ola and Majumdar, Anirudha},
  journal = {arXiv preprint arXiv:2603.09030},
  year    = {2026}
}

@article{diwa,
  title   = {DiWA: Diffusion Policy Adaptation with World Models},
  author  = {Chandra, Akshay L and Nematollahi, Iman and Huang, Chenguang and Welschehold, Tim and Burgard, Wolfram and Valada, Abhinav},
  journal = {Conference on Robot Learning (CoRL)},
  year    = 2025
}

@article{diffusionforcing,
  title={Diffusion forcing: Next-token prediction meets full-sequence diffusion},
  author={Chen, Boyuan and Mart{\'\i} Mons{\'o}, Diego and Du, Yilun and Simchowitz, Max and Tedrake, Russ and Sitzmann, Vincent},
  journal={Advances in Neural Information Processing Systems},
  volume={37},
  pages={24081--24125},
  year={2024}
}

@article{selfforcing,
  title={Self forcing: Bridging the train-test gap in autoregressive video diffusion},
  author={Huang, Xun and Li, Zhengqi and He, Guande and Zhou, Mingyuan and Shechtman, Eli},
  journal={Advances in Neural Information Processing Systems},
  volume={38},
  pages={167283--167308},
  year={2026}
}

@article{weaver,
  title={WEAVER, Better, Faster, Longer: An Effective World Model for Robotic Manipulation},
  author={Jain, Arnav Kumar and Wu, Yilin and Farebrother, Jesse and Swamy, Gokul and Bajcsy, Andrea},
  journal={arXiv preprint arXiv:2606.13672},
  year={2026}
}

@inproceedings{sarm,
  title={Sarm: Stage-aware reward modeling for long horizon robot manipulation},
  author={Chen, Qianzhong and Yu, Justin and Schwager, Mac and Abbeel, Pieter and Shentu, Fred and Wu, Philipp},
  booktitle={International Conference on Learning Representations},
  volume={2026},
  pages={17091--17116},
  year={2026}
}

@article{sarm2,
  title={SARM2: Multi-Task Stage Aware Reward Modeling for Self Improving Robotic Manipulation},
  author={Chen, Qianzhong and Zheng, Hau and Yu, Justin and Huang, Suning and Sun, Jiankai and Goldberg, Ken and Wen, Chuan and Abbeel, Pieter and Shentu, Yide and Wu, Philipp and others},
  journal={arXiv preprint arXiv:2606.10305},
  year={2026}
}

@article{3dgs,
  title={3d gaussian splatting for real-time radiance field rendering.},
  author={Kerbl, Bernhard and Kopanas, Georgios and Leimk{\"u}hler, Thomas and Drettakis, George and others},
  journal={ACM Trans. Graph.},
  volume={42},
  number={4},
  pages={139--1},
  year={2023}
}

@article{ycb,
  title={Benchmarking in Manipulation Research: The {YCB} Object and Model Set and Benchmarking Protocols},
  author={Calli, Berk and Walsman, Aaron and Singh, Arjun and Srinivasa, Siddhartha and Abbeel, Pieter and Dollar, Aaron M.},
  journal={IEEE Robotics \& Automation Magazine},
  volume={22},
  number={3},
  pages={36--52},
  year={2015}
}

@inproceedings{sam2,
  title={{SAM 2}: Segment Anything in Images and Videos},
  author={Ravi, Nikhila and Gabeur, Valentin and Hu, Yuan-Ting and Hu, Ronghang and Ryali, Chaitanya and Ma, Tengyu and Khedr, Haitham and R{\"a}dle, Roman and Rolland, Chloe and Gustafson, Laura and Mintun, Eric and Pan, Junting and Alwala, Kalyan Vasudev and Carion, Nicolas and Wu, Chao-Yuan and Girshick, Ross and Doll{\'a}r, Piotr and Feichtenhofer, Christoph},
  booktitle={International Conference on Learning Representations},
  year={2025}
}

@inproceedings{groundingdino,
  title={Grounding {DINO}: Marrying {DINO} with Grounded Pre-Training for Open-Set Object Detection},
  author={Liu, Shilong and Zeng, Zhaoyang and Ren, Tianhe and Li, Feng and Zhang, Hao and Yang, Jie and Jiang, Qing and Li, Chunyuan and Yang, Jianwei and Su, Hang and Zhu, Jun and Zhang, Lei},
  booktitle={European Conference on Computer Vision},
  pages={38--55},
  year={2024}
}

@inproceedings{sam3d,
  title={Sam 3d: 3dfy anything in images},
  author={Chen, Xingyu and Chu, Fu-Jen and Gleize, Pierre and Liang, Kevin J and Sax, Alexander and Tang, Hao and Wang, Weiyao and Guo, Michelle and Hardin, Thibaut and Li, Xiang and others},
  booktitle={Proceedings of the IEEE/CVF Conference on Computer Vision and Pattern Recognition},
  pages={7220--7232},
  year={2026}
}

@inproceedings{mujoco,
  title={Mujoco: A physics engine for model-based control},
  author={Todorov, Emanuel and Erez, Tom and Tassa, Yuval},
  booktitle={2012 IEEE/RSJ international conference on intelligent robots and systems},
  pages={5026--5033},
  year={2012},
  organization={IEEE}
}

@article{sonic,
  title={Sonic: Supersizing motion tracking for natural humanoid whole-body control},
  author={Luo, Zhengyi and Yuan, Ye and Wang, Tingwu and Li, Chenran and Casta{\~n}eda, Fernando and Chen, Sirui and Cao, Zi-Ang and Li, Jiefeng and Minor, David and Ben, Qingwei and others},
  journal={Science Robotics},
  volume={11},
  number={117},
  pages={eaed4592},
  year={2026},
  publisher={American Association for the Advancement of Science}
}

@article{twist2,
  title={Twist2: Scalable, portable, and holistic humanoid data collection system},
  author={Ze, Yanjie and Zhao, Siheng and Wang, Weizhuo and Kanazawa, Angjoo and Duan, Rocky and Abbeel, Pieter and Shi, Guanya and Wu, Jiajun and Liu, C Karen},
  journal={arXiv preprint arXiv:2511.02832},
  year={2025}
}

@article{legs,
  title={LEGS: Fine-Tuning Teleop-Free VLAs for Humanoid Loco-manipulation in an Embodied Gaussian Splatting World},
  author={Kim, Hojune and Chen, Timothy and Sun, Jiankai and Osterberg, Lars W and Chen, Qianzhong and Wang, Ke and Schwager, Mac},
  journal={arXiv preprint arXiv:2606.01458},
  year={2026}
}

@misc{zed2i,
  author={{Stereolabs}},
  title={{ZED 2i Datasheet}},
  note={Version 1.2. Accessed: Sep. 15, 2026},
  url={https://support.stereolabs.com/hc/en-us/articles/27901486677399-Where-can-I-find-the-datasheet-of-my-device}
}

@article{worldvlaloop,
  title={World-VLA-Loop: Closed-Loop Learning of Video World Model and VLA Policy},
  author={Liu, Xiaokang and Bai, Zechen and Ci, Hai and Ma, Kevin Yuchen and Shou, Mike Zheng},
  journal={arXiv preprint arXiv:2602.06508},
  year={2026}
}

@article{vlk,
  title={VLK: Learning Humanoid Loco-Manipulation from Synthetic Interactions in Reconstructed Scenes},
  author={Wang, Yen-Jen and Li, Jiaman and Chen, Sirui and Truong, Takara E and Xu, Pei and Abbeel, Pieter and Duan, Rocky and Sreenath, Koushil and Kanazawa, Angjoo and Sferrazza, Carmelo and others},
  journal={arXiv preprint arXiv:2606.30645},
  year={2026}
}

@article{simeoni2025dinov3,
  title={Dinov3},
  author={Sim{\'e}oni, Oriane and Vo, Huy V and Seitzer, Maximilian and Baldassarre, Federico and Oquab, Maxime and Jose, Cijo and Khalidov, Vasil and Szafraniec, Marc and Yi, Seungeun and Ramamonjisoa, Micha{\"e}l and others},
  journal={arXiv preprint arXiv:2508.10104},
  year={2025}
}

\end{document}